\documentclass{article}
\usepackage{ijcai18}

\usepackage{times}
\usepackage{xcolor}
\usepackage{soul}
\usepackage[utf8]{inputenc}
\usepackage[small]{caption}
\usepackage{latexsym}
\usepackage{amsmath}
\usepackage{multirow}
\usepackage{url}
\usepackage{graphicx}
\usepackage{CJKutf8}

\usepackage{amsfonts}
\usepackage{CJK}
\usepackage{amsmath}
\usepackage{bm}
\usepackage{booktabs}
\usepackage{float}
\usepackage{multirow}
\usepackage{xcolor}

\usepackage{subfigure}
\usepackage{booktabs,multirow}
\usepackage{mathrsfs}
\usepackage{array}
\usepackage{enumitem}

\newcommand{\ruber}{\textsc{Ruber}}
\newcommand{\bleu}{\textsc{Bleu}}
\newcommand{\rouge}{\textsc{Rouge}}
\newcommand{\newcite}[1]{\citeauthor{#1} [\citeyear{#1}]}

\title{One ``Ruler" for All Languages: Multi-Lingual Dialogue Evaluation with Adversarial Multi-Task Learning}

\author{
Xiaowei Tong\textsuperscript{1,2},
Zhenxin Fu\textsuperscript{1},
Mingyue Shang\textsuperscript{1},
Dongyan Zhao\textsuperscript{1,2},
Rui Yan\textsuperscript{1,2}\thanks{Corresponding author: Rui Yan (ruiyan@pku.edu.cn)} \\
\textsuperscript{1}{Institute of Computer Science and Technology, Peking University, China}\\
\textsuperscript{2}{Beijing Institute of Big Data Research, China} \\
\{tongxiaowei, fuzhenxin, shangmy, zhaody, ruiyan\}@pku.edu.cn
}

\begin{document}
\maketitle

\begin{abstract}
Automatic evaluating the performance of Open-domain dialogue system is a challenging problem. Recent work in neural network-based metrics has shown promising opportunities for automatic dialogue evaluation. However, existing methods mainly focus on monolingual evaluation, in which the trained metric is not flexible enough to transfer across different languages.  To address this issue, we propose an adversarial multi-task neural metric (ADVMT) for multi-lingual dialogue evaluation, with shared feature extraction across languages. We evaluate the proposed model in two different languages. Experiments show that the adversarial multi-task neural metric achieves a high correlation with human annotation, which yields better performance than monolingual ones and various existing metrics.
\end{abstract}

\section{Introduction}
The open-domain dialogue system is of growing interest in the field of Natural Language Processing (NLP). Its central goal is communicating with humans coherently and meaningfully; it also has wide industrial applications like XiaoIce\footnote{http://www.msxiaoice.com/} from Microsoft. Significant efforts have been made in recent years, to develop large-scale non-task-oriented dialogue system \cite{serban2016building,li2016diversity,tian2017make,yao2017towards,song2018towards}. These models adopt end-to-end neural network systems to predict the next dialogue utterance by the maximum likelihood estimation (MLE), given the previous dialogue turns.

Meanwhile, previous research has developed some successful automatic evaluation metrics. For example, BLEU \cite{papineni2002bleu} and METEOR \cite{banerjee2005meteor} are proposed for machine translation. ROUGE \cite{lin2004rouge} is widely used in automatic summarization. However, when it comes to open-domain dialogue evaluation, these metrics are shown to correlate poorly with human judgments \cite{liu2016not}. Researchers have to use those word-overlap metrics as there are few alternative efficient metrics \cite{li2016diversity,yan2016learning}. Some researchers rely on the manual annotation to evaluate their models, but it is costly and time-consuming. Therefore, having an accurate automatic dialogue evaluation model is in great need.

Very recently, some efforts have been made to develop a neural network-based metric for dialogue evaluation \cite{lowe2017towards}. It learns to predict a score of a reply given its query (previous user-issued utterance) and a groundtruth reply. This method requires massive manual annotation. RUBER \cite{tao2018ruber} tries to address the cost of annotation through negative sampling and incorporating with referenced method.

However, the above methods only extract features from monolingual corpus, in which the trained metrics are not flexible enough to transfer across different language evaluation tasks simultaneously. Besides, these methods do not exploit a multi-lingual representation to enrich the features for automatic dialogue evaluation.

In this paper, we propose an adversarial multi-task learning for multi-lingual dialogue evaluation by integrating shared knowledge from multi-lingual corpora. Specifically, we regard each monolingual evaluation as a single task and propose a shared-private model under the framework of multi-task learning \cite{caruana1998multitask,ben2003exploiting}. The multi-task learning structure contains two kinds of spaces: private and shared. The private feature spaces are used to extract the language-specific properties while the shared feature spaces capture the language-invariant properties across languages. Besides, motivated by the success of adversarial learning in domain adaption \cite{ganin2016domain,bousmalis2016domain,chen2017adversarial}, we incorporate adversarial strategy with shared spaces to enhance their ability to extract the common underlying features, and avoid the shared feature spaces being contaminated by noise.

The contributions of this paper could be summarized as follows:
\begin{itemize}
\item Multi-task learning is first introduced for automatic dialogue evaluation. It extracts not only language-specific features in private spaces but also language-invariant features in shared spaces across languages.
\item An adversarial strategy is used to strengthen the ability to extract language-invariant features in shared spaces, in which a new objective function for multi-lingual dialogue evaluation is also proposed.
\end{itemize}

We evaluated adversarial multi-task neural metric (ADVMT) on both English and Chinese evaluation tasks. Experiments show that our proposed metric significantly outperforms existing automatic metrics in terms of the Pearson and Spearman correlation with human judgements, and has a boosted performance with the help of each monolingual evaluation task.

\section{Related Work}
\subsection{Automatic Evaluation Metrics}
From the machine learning perspective, automatic evaluation metrics can be divided into non-learnable and learnable approaches. Non-learnable metrics typically measure the quality of generated sentences by heuristics (manually defined equations), such as BLEU, ROUGE and Greedy Matching \cite{rus2012comparison}. As the valid reply in dialogue systems are of high diversity under a given context, these metrics are shown to correlate poorly with human judgments \cite{liu2016not} for dialogue systems.

Compared with non-learnable metrics, learnable metrics can integrate linguistic features to enhance the correlation with human judgments through supervised learning. \newcite{lowe2017towards} develops a neural network-based metric for dialogue evaluation. RUBER \cite{tao2018ruber} addresses the cost of annotation through negative sampling and incorporating with referenced method. However, these metrics are trained in monolingual corpus, which are not flexible enough to transfer across different languages. Different from the above methods, our proposed metric extracts features from multi-lingual corpus and could be applied to different language evaluation tasks simultaneously.

\subsection{Multi-task Learning with Neural Networks}
The main concept of multi-task learning \cite{caruana1998multitask} is to extract the common underlying features between related tasks and to improve the performance of each task with the help of private features and shared knowledge through parallel training. In recent years, researchers have incorporated it with recurrent neural networks (RNN) to address various NLP problems \cite{collobert2008unified,hashimoto2017joint}.

\newcite{liu2016deep} proposes a generic multi-task framework, in which different tasks can share information by an external memory and communicate by a reading/writing mechanism. Inspired by the success of multi-task learning, we regard each monolingual evaluation as a single task and propose a shared-private model under the framework of multi-task learning for multi-lingual dialogue evaluation.

\subsection{Adversarial Neural Networks}
Adversarial neural network \cite{goodfellow2014generative} includes a neural generator $G$ and a discriminator $D$, which is trained to classify real data versus generated data. Recently, the idea of adversarial networks is applied to various NLP tasks.

 \newcite{chen2016adversarial} applies adversarial deep averaging network to transfer sentiment knowledge learned from labeled English data to low-resource languages where only unlabeled data exists. \newcite{chen2017adversarial} proposes an adversarial multi-criteria learning for Chinese word segmentation by integrating shared knowledge from multiple segmentation criteria. \newcite{liu2017adversarial} introduces an adversarial multi-task learning framework, alleviating the shared and private latent feature spaces from interfering with each other. Motivated by the success of adversarial networks, under the framework of multi-task learning, we incorporate adversarial strategy with shared spaces to enhance their ability to extract language-invariant features and propose a new objective function for multi-lingual dialogue evaluation.
\begin{figure}[htb]
\center{\includegraphics[width=0.4\textwidth]{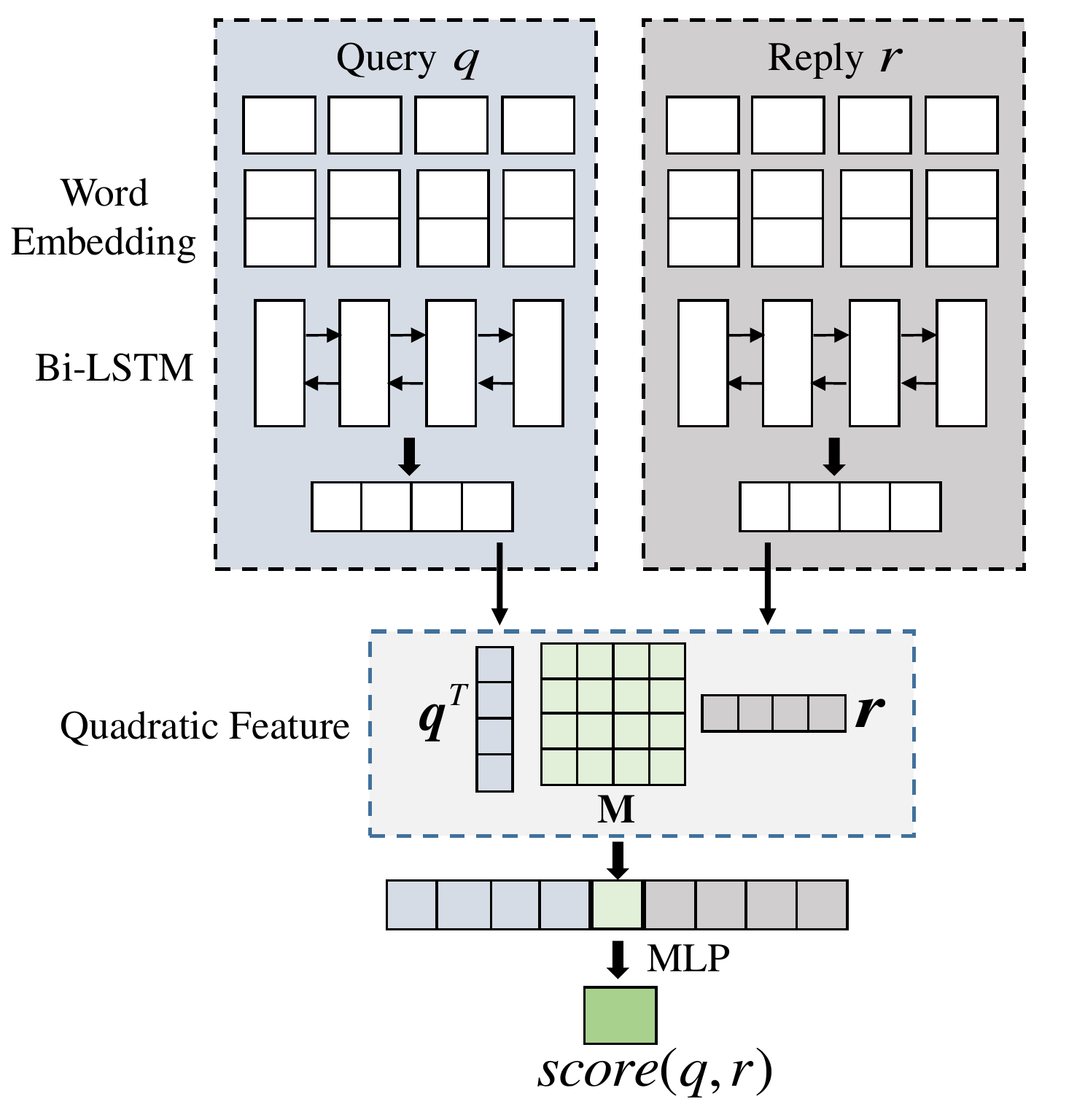}}
\caption{\label{fig:score}The single task neural metric to predict the score between a query $q$ and its reply $r$.}
\end{figure}
\section{Methodology}
Given a previous query $q$ and a reply $r$, the goal of neural network-based metric is to automatically measure the relatedness between $q$ and $r$ with a predicted $socre(q,r)$. In subsection 3.1 we introduce the neural network-based metric for monolingual dialogue evaluation, and regard it as a single task in our proposed multi-task learning framework in Subsection 3.2. In Subsection 3.3, we incorporate adversarial strategy to multi-task learning and introduce a new objective function for multi-lingual dialogue evaluation.
\begin{figure*}[htb]
\center{\includegraphics[width=1\textwidth]{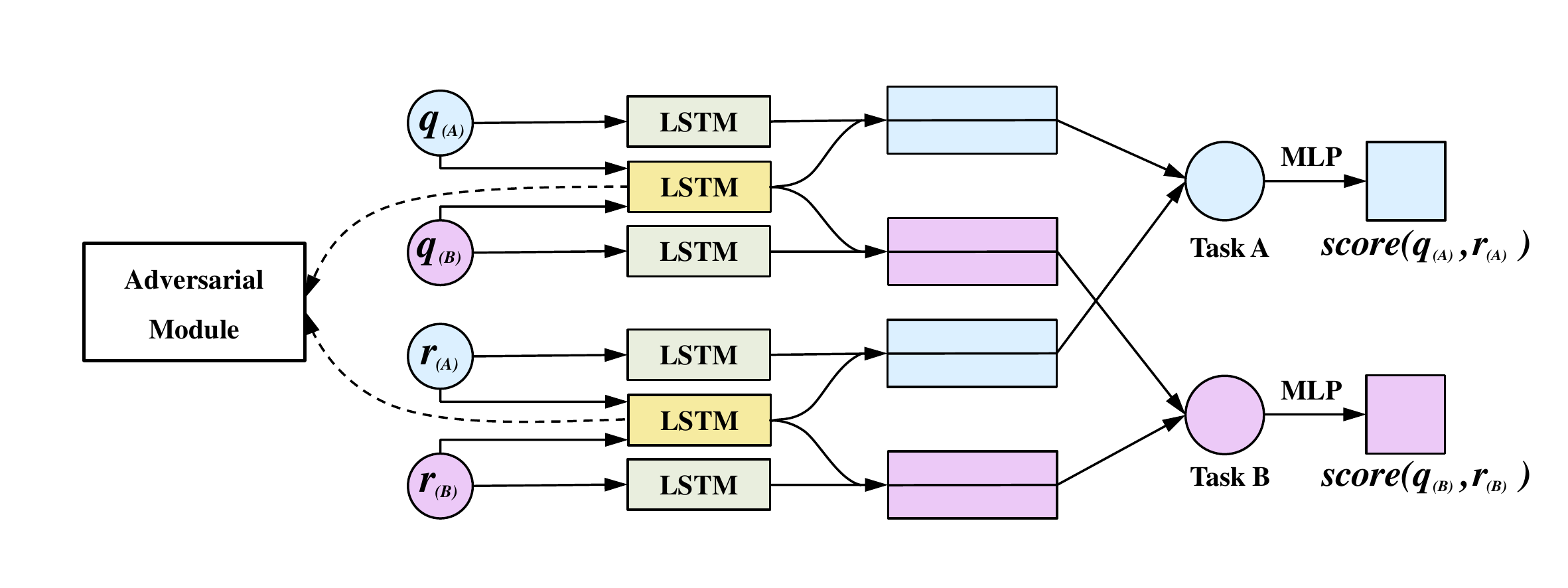}}
\caption{\label{fig:overview}Overview of adversarial multi-task neural metric for multi-lingual dialogue evaluation. The blue and purple blocks indicate different language evaluation tasks A and B, respectively. The yellow LSTM blocks are shared spaces, while the gray LSTM blocks are private spaces. }
\end{figure*}
\subsection{Neural Network-based Metric for Monolingual Dialogue Evaluation}
This subsection mainly considers a single task neural metric to predict the appropriateness of a reply with respect to a query, of which the main structure (Figure \ref{fig:score}) is inspired by \newcite{tao2018ruber}. As for each word in a query $q$ and a reply $r$, we first map them into vector representations (embedding). Then bi-directional Long Short-term Memory \cite{hochreiter1997long} (Bi-LSTM) unites with forward and backward directions are applied to capture information along the word sequence. The update of each Bi-LSTM unit can be written precisely as follows:
\begin{eqnarray}
h_t & = &\overrightarrow{h}_t \oplus \overleftarrow{h}_t \\
& = & \text{Bi-LSTM}(x_t,\overrightarrow{h}_{t-1},x_{T-t+1},\overleftarrow{h}_{t-1},\theta)
\end{eqnarray}
where \bm{$x_t$} and $T$ denote the embedding of the current input word and the last time step, and $\overrightarrow{h}_t$ is the forward hidden states. Likewise, $\overleftarrow{h}_t$ is the backward hidden states. $\oplus$ denotes the concatenation operation and all parameters in Bi-LSTM model is referred as $\theta$.
Specifically, we regard the concatenated output of both directions of Bi-LSTM, at the last time step $T$, as the representation of the whole sequence ($q$ and $r$, respectively):
\begin{eqnarray}
h_T^{(q)}=\text{Bi-LSTM}(x_T^{(q)},\overrightarrow{h}_{T-1}^{(q)},x_1^{(q)},\overleftarrow{h}_{T-1}^{(q)}, \theta^{(q)}) \\
h_T^{(r)}=\text{Bi-LSTM}(x_T^{(r)},\overrightarrow{h}_{T-1}^{(r)},x_1^{(r)},\overleftarrow{h}_{T-1}^{(r)}, \theta^{(r)})
\end{eqnarray}
after which we concatenate \bm{$q$} and \bm{$r$} to match the two utterances. In addition, we include the ``quadratic feature" that proposed in \newcite{tao2018ruber}, denoted as $\bm{q}^T\bm{Mr$}, where \bm{$M$} is a parameter matrix. Finally we use a multi-layer perceptron (MLP) to predict a scalar score of the given conversation pairs. The MLP we adopted has two layers. The $tanh$ is used as the activation function in the hidden layers of MLP, while the second layer uses $sigmoid$ to make the score bounded.

In the process of training, we consider negative sampling to avoid costly manual annotation. Negative sampling is adopted for utterance matching in previous work \cite{yan2016learning,yan2017joint} and is shown to be feasible, which could ease the burden of costly manual annotation. Concretely, given a groundtruth query-reply pair, we randomly choose another reply $r^-$ from the training set as a negative sample. The main goal of negative sampling training is to make the score of positive samples be larger than the negative samples by at least a margin $\delta$. Thus the training objective is to minimize
\begin{equation}
J_{eval}=max\{0,\delta - score(q, r) + score(q, r^-)\}\label{eval}
\end{equation}

\subsection{Multi-task Learning for Multi-lingual Dialogue Evaluation}
The method we introduced above only extract features from monolingual corpus, in which the trained metric is not flexible enough to transfer across different language evaluation tasks simultaneously and does not exploit a multi-lingual representation to enrich the features for automatic dialogue evaluation.

Inspired by the success of multi-task learning, we regard multi-lingual dialogue evaluation as multiple ``related" tasks and propose a shared-private model, which shares information across languages. This shared-private mechanism is supposed to improve the performance of each other simultaneously with the help of shared features \cite{chen2017adversarial,liu2017adversarial}.

To enable multi-task learning for multi-lingual dialogue evaluation, as depicted in Figure \ref{fig:overview}, we design two feature spaces for both tasks A and B: a private space to capture language-dependent features, and a shared space to capture language-invariant features. Each monolingual evaluation task is assigned a private Bi-LSTM layer and a shared Bi-LSTM layer. Sentences are encoded by these two kinds of Bi-LSTM layers simultaneously.

Formally, for task $k$, the query vector representations of shared layer $q_T^{(s)}$ and private layer $q_T^{(p)}$ are formed as follows:
\begin{eqnarray}
q_T^{(s)}=\text{Bi-LSTM}(x_T^{(s)},\overrightarrow{q}_{T-1}^{(s)},x_1^{(s)},\overleftarrow{q}_{T-1}^{(s)}, \theta^{(s)}) \\
q_T^{(p)}=\text{Bi-LSTM}(x_T^{(p)},\overrightarrow{q}_{T-1}^{(p)},x_1^{(p)},\overleftarrow{q}_{T-1}^{(p)}, \theta^{(p)})
\end{eqnarray}
the reply vector representation of shared layer and private layer are denoted as $r_T^{(s)}$ and $r_T^{(p)}$, likewise. As all the tasks share the shared layer, the formula of shared layer is indicated by subscript.

To compute the similarity of query-reply pair, in each monolingual evaluation task, the sentence representations from private layer and shared layer are concatenated as the final embedding. Specifically, for task $k$, the final sentence representations of query and reply are:
\begin{eqnarray}
q_{(k)}&=&q_T^{(s)} \oplus q_T^{(p)}\\
r_{(k)}&=&r_T^{(s)} \oplus r_T^{(p)}
\end{eqnarray}
which are then concatenated to calculate the $score(q_{(k)}, r_{(k)})$ for each monolingual evaluation task $k$.

\subsection{Incorporating Adversarial Strategy for Shared Spaces}
Although the shared-private model separates the feature space into shared and private spaces, there is no guarantee that sharable features do not exist in private feature space, or vice versa \cite{liu2017adversarial}. We hope that the features extracted by shared spaces is invariant across languages, under the multi-task learning framework for multi-lingual dialogue evaluation.

\begin{figure}[htb]
\center{\includegraphics[width=0.35\textwidth]{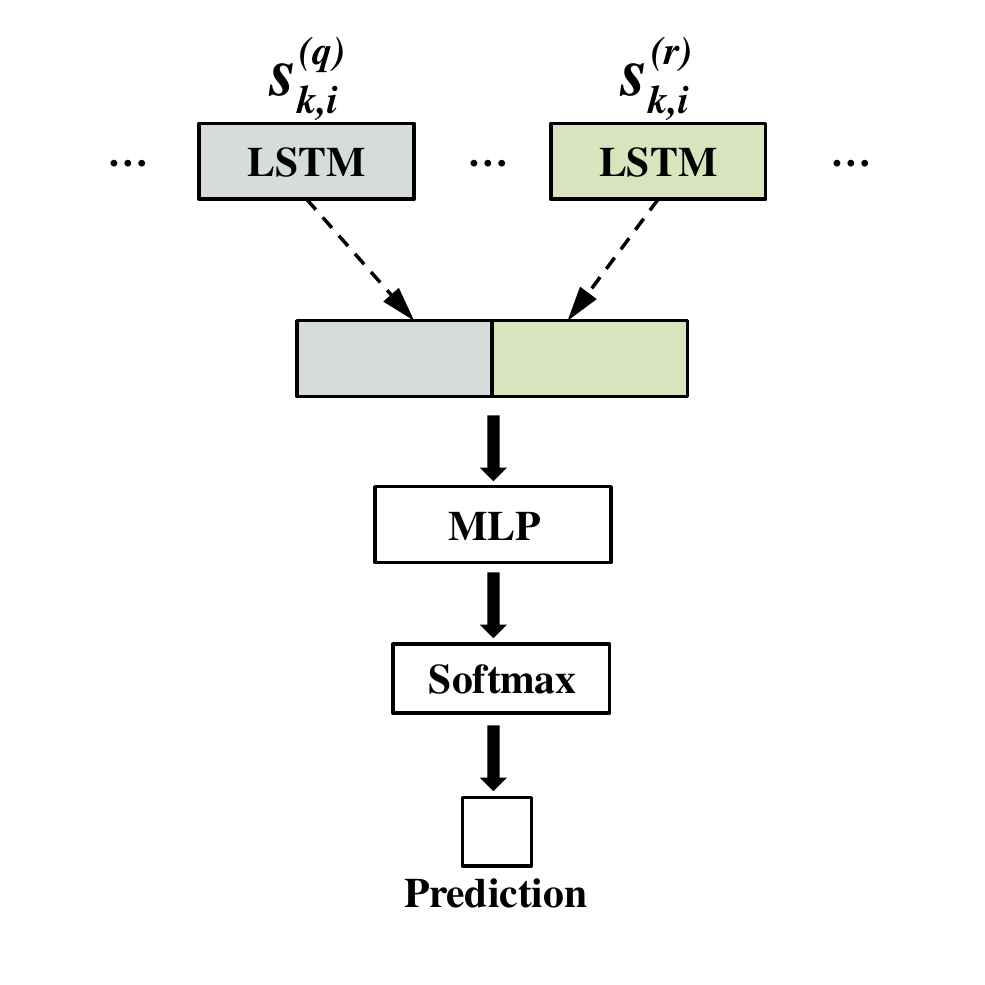}}
\caption{\label{fig:adv}Architecture of adversarial training strategy for shared spaces. The blue and green LSTM blocks are shared layers from query $s_{k,i}^{(q)}$ and reply $s_{k,i}^{(r)}$ respectively, for $i$-th query-reply pair of task $k$.}
\end{figure}

Inspired by the work on domain adaption \cite{ganin2016domain,bousmalis2016domain}, we exploit adversarial training strategy to optimize the shared layer, as shown in Figure \ref{fig:adv}. We use a discriminator to recognize which monolingual evaluation task the encoded sentence comes from. This discriminator maps the shared representation of sentences to a probability distribution, then makes a prediction of classes of monolingual evaluation tasks by its probability. The shared layers are designed to work defiantly towards a learnable multi-layer perceptron, preventing it from making an accurate prediction about the types of tasks. In this way, shared spaces are trained to be purer and less vulnerable to the contamination from private spaces.

Formally, for each monolingual evaluation task $k$, assume that there are $N_{k}$ query-reply pairs. We refer to $s_{k,i}^{(q)}$ and $s_{k,i}^{(r)}$ as shared features from query and reply respectively, for $i$-th query-reply pair of task $k$.

We further concatenate $s_{k,i}^{(q)}$ and $s_{k,i}^{(r)}$ as the input of the discriminator, denoted as $s_{k,i} = s_{k,i}^{(q)} \oplus s_{k,i}^{(r)}$. Finally, the discriminator computes the probability distribution $P(k|s_{k,i}; \Theta_D, \Theta_S)$ as:
\begin{equation}
P(k|s_{k,i}; \Theta_D, \Theta_S) = softmax(Ws_{k,i} + b)
\end{equation}
where $W$ is a learnable parameter and $b$ is a bias; $\Theta_D$ are the parameters of discriminator; $\Theta_S$ indicate the parameters of shared spaces.

Based on such adversarial structure, besides the evaluation loss $J_{eval}$, we additionally introduce an adversarial loss, so that the discriminator could help to prevent shared spaces blending with task-specific features. The adversarial loss contains two parts: one is to train the discriminator to make an accurate prediction, and the other one aims to prevent the discriminator from predicting the class of monolingual evaluation tasks.

The task discriminator learns to determine which task the feature belongs to. Thus the training objective of it is to maximize the cross entropy of predicted task distribution. The loss function is formulated as follows:
\begin{eqnarray}
J_{adv}^1(\Theta_D)=-\sum_{k=1}^K\sum_{i=1}^{N_k}logP(k|s_{k,i};\Theta_D,\Theta_S)
\end{eqnarray}
where K denotes the evaluation tasks. It updates the parameters of discriminator $\Theta_D$ to minimize the loss function.

The other part of adversarial loss aims to prevent the discriminator from predicting the class of tasks. Therefore the training objective is:
\begin{eqnarray}
J_{adv}^2(\Theta_S) = - \sum_{k=1}^K\sum_{i=1}^{N_k}-P(k|s_{k,i})logP(k|s_{k,i})
\end{eqnarray}
which is minimized by updating the parameters of shared layers $\Theta_S$. $p(k|s_{k,i};\Theta_D, \Theta_S)$ is referred as $p(k|s_{k,i})$ for short.

Combining the task evaluation loss and the adversarial loss, the final loss function of our adversarial multi-task neural metric for multi-lingual dialogue evaluation is defined as:
\begin{equation}
J = J_{eval} + J_{adv}^1 + J_{adv}^2
\end{equation}
where $J_{eval}$ is computed in Eq (\ref{eval}).

\section{Experiments}
In this section, we evaluate the correlation between our proposed metrics and manual annotation, which is the ultimate goal of automatic metrics. Our model is trained with Chinese and English datasets under the adversarial multi-task neural network framework. The overall performance is investigated on Chinese and English corpus respectively.

\subsection{Datasets}

\subsubsection{Chinese Corpus}
We build a Chinese corpus using data crawled from an online Chinese forum Douban\footnote{http://www.douban.com/}. The training set contains 1,568,241 samples, each of which consists of a query-reply pair (in text). Standard Chinese word segmentation is applied to get Chinese terms as primitive tokens. We maintain a vocabulary of 129,506 phrases ranking by the term frequency, we empirically cut the phrases that frequency is under 3.
\begin{table*}[!t] %
	\centering
	\resizebox{0.93\textwidth}{!}{
		\begin{tabular}{c|c|c|c|c|c}
			\toprule[1.5pt] \hline
			\multicolumn{2}{c|}{\multirow{2}[4]{*}{Metrics}}
			& \multicolumn{2}{c|}{\textbf{English Corpus (Twitter)}}
			& \multicolumn{2}{c}{\textbf{Chinese Corpus (Douban)}} \\
			\cline{3-6}    \multicolumn{2}{c|}{} & Pearson\scriptsize($p$-value)  & Spearman\scriptsize($p$-value)  & Pearson\scriptsize(p-value)  & Spearman\scriptsize(p-value)  \\
			\hline
			\hline
			\multirow{2}[2]{*}{Inter-annotator}
			& Human (Avg) & 0.4478{\scriptsize($<\!0.01$)} & 0.4403{\scriptsize($<\!0.01$)} & 0.4692{\scriptsize($<\!0.01$)} & 0.4708{\scriptsize($<\!0.01$)} \\
			& Human (Max) & 0.5510{\scriptsize($<\!0.01$)} & 0.5478{\scriptsize($<\!0.01$)} & 0.6068{\scriptsize($<\!0.01$)} & 0.6028{\scriptsize($<\!0.01$)} \\
			\hline
			
			\multirow{6}[2]{*}{Referenced} & \bleu-1 & 0.1214{\scriptsize($<\!0.01$)} & 0.0412{\scriptsize($<\!0.01$)} & 0.1521\scriptsize($<\!0.01$) & 0.2358{\scriptsize($<\!0.01$)} \\
			& \bleu-2 & 0.2016{\scriptsize($<\!0.01$)} & 0.2183{\scriptsize($<\!0.01$)} &\!\!\!\,-0.0006{\scriptsize($0.9914$)} & 0.0546{\scriptsize($0.3464$)} \\
			& \bleu-3 & 0.1354{\scriptsize($<\!0.01$)} & 0.1701{\scriptsize($<\!0.01$)} & \!\!\!\,-0.0576{\scriptsize($0.3205$)} & \!\!\!\,-0.0188{\scriptsize($0.7454$)} \\
			& \bleu-4 & 0.2378{\scriptsize($<\!0.01$)} & 0.1324{\scriptsize($<\!0.01$)} & \!\!\!\,-0.0604{\scriptsize($0.2971$)} & \!\!\!\,-0.0539{\scriptsize($0.3522$)} \\
			& \rouge & 0.1702{\scriptsize($<\!0.01$)} & 0.0891{\scriptsize($<\!0.01$)} & 0.1747{\scriptsize($<\!0.01$)} & 0.2522{\scriptsize($<\!0.01$)} \\
			& Greedy Matching (GM) & 0.2461{\scriptsize($<\!0.01$)} & 0.2388{\scriptsize($<\!0.01$)} & 0.3191\scriptsize($<\!0.01$) & 0.3137{\scriptsize($<\!0.01$)} \\
	   		\hline
			
			\multirow{3}[2]{*}{Unreferenced} & Single task & 0.3685\scriptsize($<\!0.01$) & 0.3702\scriptsize($<\!0.01$) & 0.4071\scriptsize($<\!0.01$) &0.4083\scriptsize($<\!0.01$) \\
			& Non-ADVMT   & 0.3823\scriptsize($<\!0.01$) & 0.3922\scriptsize($<\!0.01$) & 0.4249\scriptsize($<\!0.01$) & 0.4405\scriptsize($<\!0.01$) \\
            & \textbf{ADVMT}  & \textbf{0.3901}\scriptsize($<\!0.01$) & \textbf{0.4017}\scriptsize($<\!0.01$) & \textbf{0.4317}\scriptsize($<\!0.01$) & \textbf{0.4499}\scriptsize($<\!0.01$) \\
			\hline
			
			\multirow{4}[2]{*}{\ruber} & Min   & 0.3842\scriptsize($<\!0.01$) & 0.3721\scriptsize($<\!0.01$) & \textbf{0.4527}\scriptsize($<\!0.01$) & \textbf{0.4523}\scriptsize($<\!0.01$) \\
			& Geometric mean & \textbf{0.3928}\scriptsize($<\!0.01$) & \textbf{0.3942}\scriptsize($<\!0.01$) & 0.4523\scriptsize($<\!0.01$) & 0.4490\scriptsize($<\!0.01$) \\
			& Arithmetic mean & 0.3740\scriptsize($<\!0.01$) & 0.3688\scriptsize($<\!0.01$) & 0.4509\scriptsize($<\!0.01$) & 0.4458\scriptsize($<\!0.01$) \\
			& Max   & 0.3249\scriptsize($<\!0.01$) & 0.3126\scriptsize($<\!0.01$) & 0.3868\scriptsize($<\!0.01$) & 0.3623\scriptsize($<\!0.01$) \\
			\hline

			\multirow{4}[2]{*}{\textbf{ADVMT}+GM} & Min   & 0.4015\scriptsize($<\!0.01$) & 0.3981\scriptsize($<\!0.01$) & 0.4454\scriptsize($<\!0.01$) & 0.4535\scriptsize($<\!0.01$) \\
			& Geometric mean & \textbf{0.4267}\scriptsize($<\!0.01$) & \textbf{0.4320}\scriptsize($<\!0.01$) & \textbf{0.4698}\scriptsize($<\!0.01$) & \textbf{0.4703}\scriptsize($<\!0.01$) \\
			& Arithmetic mean & 0.3843\scriptsize($<\!0.01$ & 0.3926\scriptsize($<\!0.01$) & 0.4170\scriptsize($<\!0.01$) & 0.4214\scriptsize($<\!0.01$) \\
			& Max   & 0.2908\scriptsize($<\!0.01$) & 0.3274\scriptsize($<\!0.01$) & 0.3991\scriptsize($<\!0.01$) & 0.3999\scriptsize($<\!0.01$) \\
			\hline
			\bottomrule[1.5pt]
		\end{tabular}
	}
	\caption{Correlation between automatic metrics and human annotation. The $p$-value is a rough estimation of the probability that an uncorrelated metric produces a result that is at least as extreme as the current one; it does not indicate the degree of correlation.}
	\label{tab:result}
\end{table*}
\subsubsection{English Corpora}
We use the Twitter Corpus\footnote{http://www.twitter.com/} that contains a large number of conversations between users on the micro-blogging platform Twitter as English Corpora. The training set contains 2,537,449 query-reply pairs. Like in Chinese corpus, we maintain a vocabulary of 125,291 words, of which the frequency is higher than 5.

\subsection{Implementation Details}
\subsubsection{Hyperparameters}
For sentence encoder, we set the word embedding size $d_e$ for both tasks to 128, and they are initialized randomly and learned during training. The Bi-LSTM hidden states dimension $d_h$ is set to 256 empirically. The learning rate $\alpha$ of Bi-LSTM units is initialized to 0.001. We use a two layers multi-layer perceptron to measure the relatedness of a given query and a reply. The dimension of the first layer $d_m^1$ is set to $8 * d_h + 1$ and dimension $d_m^2$ of the second layer is 50. The optimizer used in both Bi-LSTM and MLP are Adam \cite{kingma2014adam}, and the gradient is computed by standard back-propagation. We set batch size (mini-batch) of both tasks to 128, and evaluate model on dev set after every 200 steps.
\subsubsection{Performance Evaluation}
We evaluate metrics on a generative model based on sequence-to-sequence (seq2seq) neural network \cite{bahdanau2014neural}. This generative model encodes a query into a vector representation through a recurrent neural network (RNN), and decodes this vector into a reply with another RNN. To improve the performance of seq2seq model, attention mechanism is applied.

The English and Chinese test set include 300 queries and generated replies, respectively. We had 9 volunteers to express their human satisfaction of a generated reply to a query by rating an integer score among 0, 1 and 2. Score 2 means a “good” reply, 1 borderline, and 0 bad reply.

\subsection{Results and Analysis}
Table \ref{tab:result} shows the Pearson and Spearman correlation between some metrics and human scores. The evaluated metrics are as follows.

\textbf{Referenced metrics} predict the $score(r,\hat{r})$ between the ground-truth reply $r$ and generated reply $\hat{r}$, including BLEU, ROUGE, and Greedy Matching (GM).

\textbf{Unreferenced metrics} include our Single Task, Non-adversarial Multi-task (Non-ADVMT) and Adversarial Multi-task (ADVMT) neural metrics. These metrics are unreferenced, because they predict the $score(q,\hat{r})$ between the query and its generated reply $\hat{r}$, without referring to a ground-truth reply $r$.

\textbf{RUBER} \cite{tao2018ruber} blends the referenced and unreferenced metrics by heuristics. For referenced $score(r,\hat{r})$ and unreferenced $score(q,\hat{r})$, it chooses the larger value (denoted as max), smaller value (min), and averaging (either geometric or arithmetic mean).

\textbf{ADVMT+GM} combines our Adversarial Multi-task neural metric and the Greedy Matching metric. The hybrid approach is the same as RUBER.
\begin{table*} [!ht]
\centering
    \begin{tabular}{p{4cm}<{\centering}|c<{\centering}|c|c|c|c }
    \toprule[2pt]
    \hline
    Query  & Reply &Human score& Single-task &Non-ADVMT & ADVMT \\
    \hline
    He is not very popular.& OK!!! & 0.19 & 0.30 & 0.24 &0.20\\
    \hline
    So terrible today.& Ha-ha! & 0.07 & 0.22 & 0.15 &0.08\\
    \hline
    Where are you in Baoshan? & I'm in Minhang. & 0.79 & 0.59 & 0.69 &0.73\\
    \hline
    There is a teaching video. & Oh, thank you! & 0.79 & 0.36 & 0.40 &0.60\\
    \hline
    This is not an easy job. & Is there any value for this job? & 0.36 & 0.11 & 0.17 &0.32\\
    \hline
    \bottomrule[2pt]
    \end{tabular}
\caption{\label{tbl: cases} Selected cases with query and generated reply in Chinese and English, and Chinese is translated into English here. All the scores are mapped into the same section of $[0, 1]$ for directly comparing.}
\end{table*}
\subsubsection{Overall Performance}
The first observation in table \ref{tab:result} is that our unreferenced metrics are more correlated than those referenced metrics with human judgment in both English and Chinese evaluation tasks. This is because the referenced metrics mainly capture the similarity, but the rich semantic relationship between queries and replies necessitates more complicated mechanisms like neural networks. Besides, the referenced metrics have to rely on the information of both reference reply and generated reply, while neural network-based metrics use no reference but query and model reply. This observation shows that the query alone is also informative and that negative sampling could help to train the evaluation metrics, although it does not require human annotation as labels.

In unreferenced metrics, compared to single-task trained metric and non-adversarial multi-task trained metric, the ADVMT metric trained under the framework of adversarial multi-task achieves a best result in the correlation with human judgment. Although the corpora used in multi-task training are in different languages, the shared-private architecture shows the ability to extract useful language-invariant features. Thus when evaluating on a single language dialogue system, with the help of that shared information across languages, the performance is boosted. In addition, this result shows that incorporating the adversarial strategy could strengthen the ability to extract language-invariant features in shared spaces and help to prevent the shared spaces of features from being interfered by private spaces.

We combine the referenced metric GM and unreferenced metric ADVMT, and the hybrid approach is the same as RUBER. Experiments show that ADVMT+GM metric achieves the best result than RUBER peak performance, when choosing the geometric mean blended strategy. What’s more, in both ADVMT+GM and RUBER metrics, choosing the larger value (max) is too lenient, and is slightly worse than other strategies. More importantly, our ADVMT metric is trained in multi-lingual dataset, which could be applied in multi-lingual dialogue evaluation simultaneously, while the RUBER metric should be trained in each monolingual dataset respectively and ignores massive information across languages.

\subsubsection{Case Study}
Table \ref{tbl: cases} illustrates some examples of Single Task, Non-ADVMT, ADVMT neural metric.
As for the non-universal reply, we find that our ADVMT metric tends to give a closer score with the human score than the Single Task metric and Non-ADVMT metric.
We further observe that a common problem of generative model is that it tends to generate a universal reply, such as ``Ha-ha." and ``Ok!" We observe that our ADVMT metric tends to give a lower score when it comes to such universal replies, while the Single Task metric and Non-ADVMT metric gives a relatively high score. Improving the diversity in generative model remains a challenging problem. But it may be furthered if the evaluation metrics used in training process encourage replies that of high diversities, and discourage those universal replies.

\section{Conclusion and Future Work}
In this paper, we propose an adversarial multi-task neural metric for multi-lingual dialogue evaluation, using shared feature extraction across languages. In addition, we incorporate adversarial strategy to shared spaces, which aims to guarantee the purity of shared feature spaces. Our proposed model regards models that trained in different language corpora as a single task and integrates each single task under the framework of adversarial multi-task learning. Experiments show that the proposed model outperforms the monolingual ones and various existing metrics.

An important direction of future research is evaluating the ability of the proposed metric to transfer knowledge from one language to another. There could be a problem of lacking training corpus when it comes to the dialogue system on minority languages. As the proposed metric could extract information across languages, the performance of multi-lingual evaluation metrics, which are trained on some majority languages with a massive corpus, of transferring the shared knowledge to the minority languages is worth exploring.
\section*{Acknowledgments}
We appreciate the contribution from Chongyang Tao. Besides, we would like to thank the anonymous reviewers for their constructive comments. This work was supported by the National Key Research and Development Program of China (No. 2017YFC0804001), the National Science Foundation of China (No. 61672058). Rui Yan was sponsored by CCF-Tencent Open Research Fund and MSRA Collaborative Research Program.
\bibliographystyle{named}
\bibliography{ijcai18}
\end{document}